\documentclass[conference]{IEEEtran}
\IEEEoverridecommandlockouts
\usepackage{cite}
\usepackage{amsmath,amssymb,amsfonts}
\usepackage{algorithmic}
\usepackage{graphicx}
\usepackage{textcomp}
\usepackage{xcolor}
\def\BibTeX{{\rm B\kern-.05em{\sc i\kern-.025em b}\kern-.08em
    T\kern-.1667em\lower.7ex\hbox{E}\kern-.125emX}}
\begin{document}

\title{IMSSA: Deploying modern state-space models on memristive in-memory compute hardware\\
\thanks{Identify applicable funding agency here. If none, delete this.}
}

\author{\IEEEauthorblockN{1\textsuperscript{st} Sebastian Siegel}
\IEEEauthorblockA{\textit{Peter-Gruenberg-Institute (PGI-14)} \\
\textit{Forschungszentrum Juelich GmbH}\\
Juelich, Germany \\
s.siegel@fZ-juelich.de}
\and
\IEEEauthorblockN{2\textsuperscript{nd} Ming-Jay Yang}
\IEEEauthorblockA{\textit{Peter-Gruenberg-Institute (PGI-14)} \\
\textit{Forschungszentrum Juelich GmbH}\\
Juelich, Germany \\
m.yang@fz-juelich.de}
\and
\IEEEauthorblockN{3\textsuperscript{rd} John-Paul Strachan}
\IEEEauthorblockA{\textit{Peter-Gruenberg-Institute (PGI-14)} \\
\textit{Forschungszentrum Juelich GmbH}\\
Juelich, Germany \\
j.strachan@fz-juelich.de}
}

\maketitle

\begin{abstract}
Processing long temporal sequences is a key challenge in deep learning. In recent years, Transformers have become state-of-the-art for this task, but suffer from excessive memory requirements due to the need to explicitly store the sequences. To address this issue, structured state-space sequential (S4) models recently emerged, offering a fixed memory state while still enabling the processing of very long sequence contexts. The recurrent linear update of the state in these models makes them highly efficient on modern graphics processing units (GPU) by unrolling the recurrence into a convolution. However, this approach demands significant memory and massively parallel computation, which is only available on the latest GPUs.\par
In this work, we aim to bring the power of S4 models to edge hardware by significantly reducing the size and computational demand of an S4D model through quantization-aware training, even achieving ternary weights for a simple real-world task. To this end, we extend conventional quantization-aware training to tailor it for analog in-memory compute hardware. We then demonstrate the deployment of recurrent S4D kernels on memrisitve crossbar arrays, enabling their computation in an in-memory compute fashion. To our knowledge, this is the first implementation of S4 kernels on in-memory compute hardware.

\end{abstract}

\begin{IEEEkeywords}
state-space models, in-memory computing, memristive crossbar arrays
\end{IEEEkeywords}

\section{Introduction}
Processing long temporal sequences presents a significant challenge for many deep learning algorithms \cite{lstmp, rnnlongmemory}. Transformers \cite{transformer} address this by explicitly storing projections of a long history of input tokens \cite{attention}. However, this dynamic allocation of memory, which scales quadratically with the sequence length \cite{lra}, results in substantial memory requirements. Consequently, these algorithms become impractical for resource-constrained applications, such as natural language processing in off-grid environments or remote sensor data processing. \par
A promising solution to this issue is the emergence of state-space sequential models, such as S4(D) \cite{gu2022s4d} and MAMBA\cite{gu2023mamba}. They overcome the training challenges associated with classical recurrent neural networks by employing a linear state transition, which can be unrolled into a convolutional kernel. This allows for efficient training and execution on modern GPUs, which have the memory capacity to store the entire convolutional kernel.  However this approach is impractical on edge computing hardware, where memory is limited. An alternative is to perform the state update in a recurrent manner, which reduces memory requirements but necessitates an efficient method for computing vector-matrix multiplications (VMM). \par
An efficient choice for such operations are memristive crossbar arrays (MCBA)  \cite{memCBA1, memCBA2, memCBA3, memCBA4, memCBA5, memCBA6, memCBA7} which offer analog in-memory computing of VMMs. MCBAs use emerging non-volatile resistive switching devices and have been used for various VMM accelerators, because they allow computing a VMM in a single operation \cite{VMM1, VMM2, VMM3, VMM4}. However, to our knowledge, their application to modern state-space models has not been investigated so far, which is the aim of this work. \par
In this manuscript, we first illustrate the hardware-aware training process required for successfully deploying the recurrent kernels of an S4D model on an MCBA. To address this, we incorporate the limited dynamic range of memristive devices into a quantization-aware training approach. We also introduce the In-Memory State-Space model Accelerator architecture (IMSSA), which maps an entire S4D kernel onto a single MCBA, and we demonstrate its performance on a simple real-world task. Finally, we highlight the critical role of quantization in enabling the effective deployment of state-space kernels onto noisy computational substrates.

\section{Methods}
\subsection{The S4(D) model}
At the core of the S4(D) model are the so-called HiPPO kernels, originally proposed by Gu et al. \cite{gu2020hippo}. These kernels use a vector $\textbf{B} \in \mathbb{R}^{N}$ to project a one-dimensional input signal $\text{u(t)}$ into the higher dimension of the state $\text{x(t)} \in \mathbb{R}^{N}$. The state is recurrently updated via the projection matrix $\mathbf{A} \in \mathbb{R}^{NxN}$ and the output of each kernel is calculated by multiplying the state with a vector $\textbf{C} \in \mathbb{R}^{N}$. While it is possible to include a skip connection directly from the input to the output, this is not implemented in the current work. The complete kernel can be formulated as
\begin{equation}
    \begin{aligned}
        \frac{d x(t)}{dt} &= \mathbf{A}x(t) + \mathbf{B}u(t) \\
        y(t) &= \mathbf{C}x(t) .
    \end{aligned}
    \label{eq:kernel_continuous}
\end{equation}
In time-discrete systems, for example for training on conventional GPUs, the kernel must be discretized in time. A commonly used method, which is also applied in this work, is the zero-order hold method. For a constant time-step $\Delta$, which is a trainable parameter, this approach results in new time-discrete kernels 
\begin{equation}
    \begin{aligned}
        x_{t} &= \overline{\mathbf{A}}x_{t-1} + \overline{\mathbf{B}}u_t\\
        y_t &= \overline{\mathbf{C}}x_t .
    \end{aligned}
    \label{eq:kernel_discrete}
\end{equation}
An S4 layer consists of $\text{H}$ kernels running in parallel, combined with a linear mixing layer. The mixing layer projects the outputs of the HiPPO kernels back to the input dimension $\text{H}$ of the next layer. Most S4 models are composed of multiple such layers stacked in series, typically enclosed by linear encoder and decoder layers.\par
Since these models are trained on conventional GPUs, the recurrent vector-matrix multiplication imposes a computational bottleneck. In addition to unrolling the recurrence into a convolution, an important step is the diagonalization of the matrix $\textbf{A}$. This results in a complex matrix, which in turn leads to a complex state $\text{x(t)}$ and a complex matrix $\textbf{C}$. Models utilizing this approach are referred to as S4D models. \par

\subsection{Quantization-aware training}
For deployment on low-resolution inference hardware, the parameters of the S4D model typically need to be stored and computation executed using quantized integer formats. However, the model is trained on high-precision floating point GPUs. While post-training quantization of the high-precision model is possible, it significantly reduces accuracy, even at moderate quantization levels. To achieve more aggressive quantization, a common technique is quantization-aware training (QAT). Abreu et al. \cite{abreu2024qs5} have demonstrated the benefits with this approach on a similar state-space model (S5). \par
A common approach to QAT in auto-differentiating machine learning frameworks is the use of the straight-through estimator. In this method, the forward pass of the model employs quantized parameters and activations, while the backward pass and gradient updates are performed with full precision. This approach has been successfully applied to numerous recent QAT tasks \cite{abreu2024qs5, ma2024158bit}. As an extension of this method, we introduce the training for a specific constant dynamic range by choosing a constant $f_{scale}$ in the quantization function
\begin{equation}
    x_{i, quant} = round\left(x_i * \frac{n_{levels}}{f_{scale}}\right) * \frac{f_{scale}}{n_{levels}} .
    \label{eq:quantization}
\end{equation}

\subsection{Task description}
The target application of the S4D model discussed here involves classification tasks at the edge, such as key word spotting. To evaluate its performance, we tested the model on a two-class subset of the Heidelberg Digits raw audio dataset. Specifically, the model is trained to distinguish between vocal utterances of the English words "zero" and "one". The audio files are normalized and downsampled by a factor of 64, resulting in 871 samples per input.

\subsection{Memristive crossbar arrays}
\begin{figure}[th]
\includegraphics[scale=1]{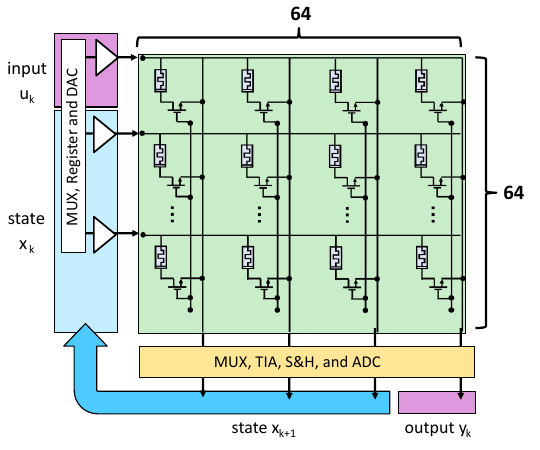}
\caption{Memristve State-Space Model Accelerator (IMSSA) kernel implementation on an MCBA.}
\label{fig:MCBA_implementation}
\end{figure}

Memristive devices are an emerging class of non-volatile memory elements, functioning as two-terminal resistors with electronically adjustable conductance. Their conductance can be programmed gradually, enabling their use in constant time VMM through Ohm's Law and Kirchhoff's Law. To achieve this, memristive devices are typically arranged in matrix-like structures (see Figure \ref{fig:MCBA_implementation}) to perform the multiplication of an applied voltage vector with a stored conductance matrix in a single operation, in contrast to the quadratically scaling complexity of this operation on classical digital hardware. Memristive devices are often put in series with a transistor for more precise programming. Once programmed, these devices behave like passive resistors and can only represent real-valued positive conductances. However, S4D kernels are complex-valued and can have both positive and negative real and imaginary components. As a result, each element of the kernels must first be expanded similar to \cite{zhao2023} into its real and imaginary parts for further processing,
\begin{equation}
    \underline{m} * \underline{v} = 
    \begin{pmatrix}
        m_r & -m_i \\
        m_i & m_r
    \end{pmatrix} *
    \begin{pmatrix}
        v_r \\
        v_i
    \end{pmatrix}
    =
    \begin{pmatrix}
        i_r \\
        i_i
    \end{pmatrix}
    = \underline{i}
\end{equation}
 and then by positive and negative parts through

\begin{equation}
    g * v = 
    \begin{pmatrix}
        g^+ & g^-\\
        g^- & g^+
    \end{pmatrix} * 
    \begin{pmatrix}
        v^+\\
        v^-
    \end{pmatrix} =
    \begin{pmatrix}
        i^+\\
        i^-
    \end{pmatrix} .
    \label{eq:neg_exp}
\end{equation}

Eventually, each matrix element in the mathematical kernel yields a 4x4 matrix of memristive conductances and inputs and outputs are expanded into four element vectors following

\begin{equation}
    \begin{pmatrix}
        g_r^+ & g_r^- & g_i^- & g_i^+ \\
        g_r^- & g_r^+ & g_i^+ & g_i^- \\
        g_i^+ & g_i^- & g_r^+ & g_r^- \\
        g_i^- & g_i^+ & g_r^- & g_r^+ 
    \end{pmatrix} * 
    \begin{pmatrix}
        v_r^+\\
        v_r^-\\
        v_i^+\\
        v_i^-
    \end{pmatrix} = 
    \begin{pmatrix}
        i_r^+\\
        i_r^-\\
        i_i^+\\
        i_i^-
    \end{pmatrix} .
\label{eq:complex_to_mem}
\end{equation}

\section{Results and Discussion}

\subsection{Hardware-aware training for aggressive quantization}
In conventional QAT, the scaling factor $f_{\text{scale}}$ in equation \ref{eq:quantization} is specific to each parameter and typically depends on either the maximum or the mean of that parameter. For S4D models, this implies that the real and the imaginary part of the complex matrix $\mathbf{A}$ will have different $f_{\text{scale}}$ values, leading to different quantization levels for each part. Consequently, the kernel function in \ref{eq:kernel_continuous} must account for different dynamic ranges of the parameters. In the case of complex multiplication, parameters must be further normalized to ensure that summands with different quantization levels can be properly added. This issue becomes particularly critical when deploying the model on analog computing hardware, where signal dynamic range is typically constrained by supply voltages. Separating signals with different dynamic ranges would require physical separation, along with analog multiplication when the signals are combined. Both of which result in significant power and silicon area overheads. \par
To address this issue, we introduce quantization-aware training with a fixed dynamic range. This is implemented in the training algorithm by setting $f_{\text{scale}} = const.$, ensuring that the real and imaginary part of the $\mathbf{A}$ matrix share the same dynamic range. As a result, normalization between real and imaginary computations is no longer required. The impact of this method on model performance for a simple real-world audio classification task is negligible, with the model achieving a classification accuracy of $95.06 \%$, comparable to the model without fixed dynamic range. \par
\begin{figure}[th]
    \centering
    \includegraphics{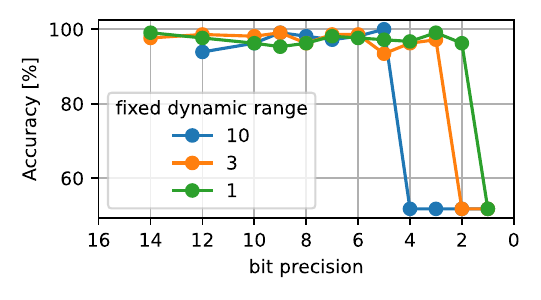}
    \caption{Training the S4D model with a smaller fixed dynamic range allows for quantization-aware training to lower bit precision without loss of accuracy.}
    \label{fig:quant_dev}
\end{figure}
Figure \ref{fig:quant_dev} illustrates that various predefined fixed dynamic ranges for the $\mathbf{A}$ matrix can be applied during training. Notably, a smaller fixed dynamic range allows the model to be trained with more aggressive quantization. For instance, with a fixed dynamic range of $1$, the model can achieve near-baseline performance even with a  2 bits quantization of $\mathbf{A}$ matrix, which is not feasible without a fixed dynamic range or larger ranges such as $3$ or $10$. Since quantization normalizes each parameter to a symmetric range around zero, and the real part of the $\mathbf{A}$ matrix is negative or zero while the imaginary part is either positive or zero, a 2 bits quantization effectively reduce the $\mathbf{A}$ matrix to ternary values.\par
It should be noted that for certain artificial benchmark tasks, such as the sequential CIFAR10 dataset, this method requires higher bit precision compared to using a dynamic and individual $f_{scale}$ value.

\subsection{IMSSA: Realizing the S4D kernel in a single memristive crossbar array}
\begin{figure}
    \includegraphics{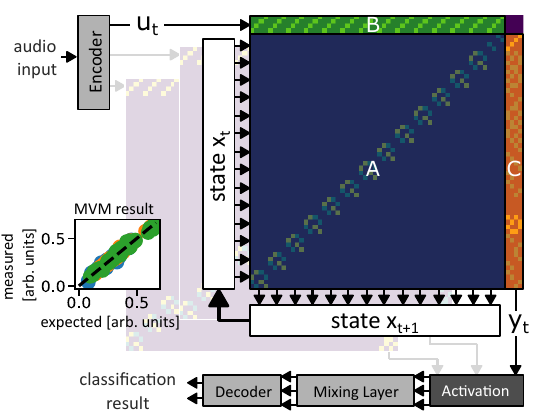}
    \caption{Integration of the IMSSA architecture in the S4 model. Memristive conductances are color-coded with the respective matrices as overlays.}
    \label{fig:messa}
\end{figure}

The core computation of S4D models involves the recurrent computation of the state in the kernel $\textbf{A}x$, while incorporating the input $\textbf{B}u$ and extracting the output $y=\textbf{C}x$. These three operations consist of vector or (diagonal) matrix multiplications, meaning that the computational effort scales linearly with the dimension of the state. Instead of treating the three operations separately, which would require  physically separating them into different tiles and broadcasting signals between them, we propose to combine all three in a single memristive crossbar array to fully leverage the benefits of in-memory computing. Consequently, the trained kernel matrices $\textbf{A}$, $\textbf{C}$, and the (in this case) unit vector $\textbf{B} = \textbf{1}$ are programmed into a memristive crossbar array as illustrated in Figure \ref{fig:messa}. The complex values of these matrices are expanded into memristive conductances using the previously described method. For the diagonal $\textbf{A}$ matrix, the resulting 4x4 blocks are visible, while the vector $\textbf{B}$ is arranged horizontally with small diagonals, and the vector $\textbf{C}$ appears as a vertical block four devices wide. The input signal $\text{u}$ is applied in a complex expanded form across the first four rows. In the subsequent rows, a vector of voltage signals representing the state $\text{x}$ is applied. At each time step, this vector is obtained by converting the output currents of the MCBA from the first $N*4$ columns for the next time step, creating a time-delayed recurrent connection. The last four output currents represent the result of the output function $\textbf{C}x$. As a result, this implementation solves the kernel computation as follows
\begin{equation}
    \begin{aligned}
        x_{t} &= \overline{\mathbf{A}}x_{t-1} + \overline{\mathbf{B}}u_t\\
        y_{t-1} &= \overline{\mathbf{C}}x_{t-1}
    \end{aligned}
    \label{eq:kernel_MCBA}
\end{equation}
executing all operations in an in-memory compute fashion in a single operation. The memristive kernel \ref{eq:kernel_MCBA}  represents only a slight deviation from the original time-discrete kernel \ref{eq:kernel_discrete}, namely a time-shift of the output of one time step.\par
From this approach for deploying the S4D kernel on an MCBA array, one cane deduce the maximum dimension of the state, which serves as a hyper-parameter for neural network training. The MCBA accelerator chip \cite{b8, b9, b10} we use for deployment features three MCBAs, each with a size of 64x64 memristive devices. Given that four rows and four columns are reserved for the horizontal $\textbf{B}$ and the vertical $\textbf{C}$ vector, this leaves a 60x60 sub-array available for potential implementation, in which a 15x15 diagonal matrix \textbf{A} could be realized. However, for error correction purposes, we opt to use a 14x14 matrix instead. Since we want to deploy a model on a single chip, the number of layers is set to one , and the number of parallel kernels is limited to three, corresponding to the number of MCBAs per chip. Using the conversion equation \ref{eq:complex_to_mem}, and following the QAT process described above, the kernels are written on the MCBAs with memristive conductance values ranging from $7\mu S$ to $200 \mu S$. The resulting memristive arrays are shown in Figure \ref{fig:messa}. With these kernels, the model achieves a classification accuracy of $81.69 \%$ on the test data set. While this is significantly above random guessing at $50\%$, it represents a notable drop compared to the software model's accuracy of $95.06 \%$. As illustrated in Figure \ref{fig:noise_impact}, this performance is slightly below the distribution of software-calculated accuracies for the write noise level found during programming. The discrepancy can be attributed to faulty devices within the MCBAs, which remain  stuck in a high conductive state, causing large incorrect activations. Such issues have been previously reported in the literature for this accelerator platform \cite{b9}. Without these extreme programming errors, typically occurring in one to three devices per kernel, the software model accuracy of $95.61\%$ could be fully restored. \par
To the best of our knowledge, this result presents the first demonstration of an S4D kernel implemented on an analog in-memory compute substrate.


\subsection{Write-noise resilience through quantization}
Memristive crossbar architectures today can be realized using various materials and physical mechanisms \cite{memCBA1, memCBA2, memCBA3, memCBA4, memCBA5, memCBA6, memCBA7}. However, most of these approaches suffer from inherent write noise or require complex write schemes to achieve high programming accuracies\cite{memProg1, memProg2, memProg3, memProg4, memProg5, memProg6, memProg7}. Even though the kernel is typically programmed once during network initialization, a practical use case would also involves reprogramming for updates or newer versions of the network. This imposes constraints on both time and energy budget for programming cycles, leading to some unavoidable level of write noise.\par
However, strong quantization of the network can mitigate the impact of write noise. To evaluate this, we simulate the deployment of the kernels as described above by injecting Gaussian write noise with zero mean and varying $\sigma$ into the trained quantized kernels. We then investigate the resulting inference accuracy on the test data set. 
\begin{figure}[th]
    \centering
    \includegraphics{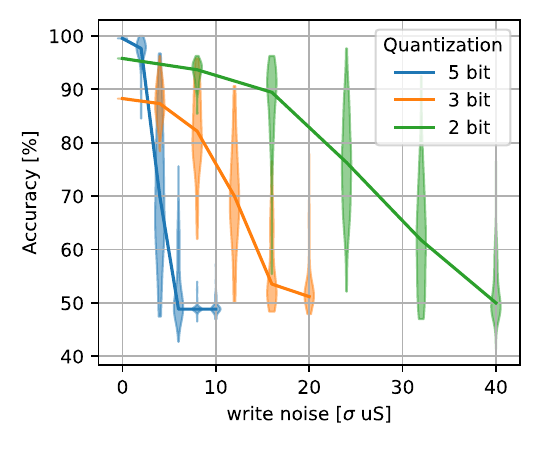}
    \caption{Influence of write noise on model performance for different kernel quantization. Violin plots indicate the distribution of 100 instantiations around the median.}
    \label{fig:noise_impact}
\end{figure}
Figure \ref{fig:noise_impact} shows that for a quantization of the kernel parameters to 5 bits already $\sigma = 5 \mu S$ lead to a strong decrease in classification accuracy. For a 2 bit quantization, more than $\sigma = 15 \mu S$ can be sustained without significantly reducing the performance. This shows how a strong quantization can make a network robust enough to yield high performance on a noisy substrate. 

\section{Conclusion}
In this work, we demonstrate a comprehensive hardware-aware training process for deploying an S4D model onto an analog in-memory computing substrate. The approach can be generalized to other state-space model architectures like S4 or S5 and other in-memory compute architectures. Using memristive crossbar arrays as an example, we account not only for the limited programming precision of memristive devices, but also for the constrained dynamic range of signals in an analog computing system. By extending a common quantization-aware training method with a fixed dynamic parameter range, we successfully train and deploy an S4D network for an audio classification task. Furthermore, we explore how quantization helps mitigate the effects of programming noise on the network performance. To the best of our knowledge, this is the first dedicated hardware implementation of modern state models using in-memory and analog computing principles, which is an important step toward the efficient deployment of these models in edge computing applications.

\subsubsection*{Achnowledgements}
This work is supported by the NeuroSys project as part of the initiative “Clusters4Future” funded by the Federal Ministry of Education and Research BMBF (grant no. 03ZU1106CB).

\vspace{12pt}

\end{document}